\documentclass[lettersize,journal]{IEEEtran}
\usepackage{amsmath,amsfonts}
\usepackage{algorithmic}
\usepackage{algorithm}
\usepackage{array}
\usepackage[caption=false,font=footnotesize,labelfont=rm,textfont=rm]{subfig}
\usepackage[T1]{fontenc}
\usepackage{aecompl}
\usepackage{textcomp}
\usepackage{stfloats}
\usepackage{url}
\usepackage{verbatim}
\usepackage{graphicx}
\usepackage{cite}
\usepackage{bbm}
\usepackage{dsfont}

\begin{document}

\title{Graph-Based Safe  Reinforcement Learning for Multi-Agent Systems with Time-Varying Topology}

\author{
\IEEEauthorblockN{Xiao Sizhe$^{a}$, Dong Lijing$^{a*}$, Bai Rui$^a$, Tan Xin$^{a}$}

\IEEEauthorblockA{$^a$ School of Mechanical, Electronic and Control Engineering, Beijing Jiaotong University, Beijing, China}
}

\markboth{Journal of \LaTeX\ Class Files,~Vol.~14, No.~8, August~2021}%
{Shell \MakeLowercase{\textit{et al.}}: A Sample Article Using IEEEtran.cls for IEEE Journals}


\maketitle
\pagestyle{empty}
\thispagestyle{empty} 
\begin{abstract}
This paper presents a graph-based safe multi-agent reinforcement learning (MARL) framework for cooperative navigation with time-varying topology. To address the critical challenge of ensuring safety in environments with sensing constraints, a safety-decoupled mechanism is introduced through a Control Barrier-Like Function (CBLF) action screening layer. This mechanism bridges the gap between discrete LiDAR perception and continuous safety constraints, ensuring that physical safety constraints are strictly satisfied regardless of the learning progress. Building upon this safety foundation, a unified structural architecture is proposed, integrating a attention-based actor and a Graph Attention Network (GAT) centralized critic. The actor utilizes a value vector reconstruction mechanism that explicitly encodes relative geometric relations through a collaborative tracking error matrix, enabling scale-insensitive policy learning under time-varying communication topologies. Meanwhile, the GAT-based critic models evolving interaction structures for accurate global value estimation. The proposed framework is validated on real differential-drive robot platforms, and experimental results demonstrate superior stability and safety in dynamic scenarios with limited fields-of-view.
\end{abstract}


\section{Introduction}
\IEEEPARstart{M}{ulti-agent} cooperative navigation constitutes a fundamental problem in robotics, defined as the process where a group of autonomous agents coordinates their movements to reach designated target locations safely while avoiding collisions with obstacles and peers\cite{1}. This capability forms the operational basis for various real-world applications, including automated logistics, swarm robotics, and unmanned vehicle formations. To address these problems, existing methods have evolved from traditional rule-based algorithms, such as Whale Optimization Algorithm\cite{2} and Particle Swarm Optimization\cite{3}, to learning-based algorithms. 

In recent years, MARL has been widely adopted in cooperative navigation, enabling agents to acquire decentralized control policies through extensive interactions with the environment\cite{4},\cite{5},\cite{6}. While these algorithms have achieved notable performance in several control tasks by optimizing reward functions and demonstrated efficacy in simulated environments, deploying such methods in physical constraint systems still faces significant challenges. Particularly, handling dynamic coordination structures and providing strict safety guarantees remain difficult, as relying solely on reward penalties for collision avoidance fails to ensure physical safety.

Furthermore, existing safe MARL approaches\cite{7},\cite{8} frequently assume perfect global state information or rely on abstract state-space constraints, which are difficult to map onto actual sensor capabilities. Under realistic sensing constraints, these algorithms may generate control inputs that are geometrically unverifiable and hazardous. To bridge the gap between abstract safety and physical executability, raw distance measurements from LiDAR sensors provide a direct, unmediated geometric representation of the local environment, allowing for rigorous spatial occupancy verification. This limitation necessitates a mechanism that decouples physical safety constraints from the reinforcement learning objective by restricting actions within a geometrically feasible safe region. Furthermore, existing MARL architectures\cite{9},\cite{10},\cite{11} predominantly utilize standard multi-layer perceptrons in both actor and critic networks, which lack the structural modeling capability required for dynamic multi-agent environments. As the adjacency topology varies and the number of observable neighbors fluctuates, these traditional models struggle to accurately encode relative geometric relationships and estimate global values from limited local observations. Consequently, developing a unified collaborative learning framework that simultaneously addresses time-varying graph structures and decoupled physical safety verification remains a primary challenge for deploying multi-agent systems.

To address the aforementioned challenges, a graph-based safe multi-agent reinforcement learning framework is proposed for cooperative navigation with time-varying topology. The main contributions of this work are threefold:
\begin{itemize}
\item[$\bullet$] \textbf{Geometrically verifiable action screening via LiDAR-
based CBLF}. This paper introduces the CBLF, a discrete formulation of the traditional CBF, explicitly designed to leverage discrete sensor information. By utilizing this discrete geometric representation, this paper introduce a safety-decoupled action screening mechanism that projects policy outputs onto a feasible safe region at each time step.
\item[$\bullet$] \textbf{Graph-based collaborative learning architecture}. This paper develops an integrated framework featuring an attention-based actor and a GAT-based centralized critic. The actor explicitly encodes relative geometric relations through a value vector reconstruction mechanism, while the critic dynamically models evolving interaction topologies to achieve accurate global value estimation and scale-insensitive policy learning.
\item[$\bullet$] \textbf{Real-world deployment and Validation}. The proposed framework is successfully deployed on physical differential-drive robot platforms. This empirical validation demonstrates the framework's engineering feasibility, cooperative stability under dynamic topologies, and robust safety in scenarios.
\end{itemize}


The remainder of this article is structured as follows. Section II reviews the related work on multi-agent reinforcement learning and safe navigation. Section III provides the problem formulation for cooperative navigation under dynamic topologies and physical constraints. Section IV details the proposed graph-based safe multi-agent reinforcement learning algorithm, comprising the CBLF-based action screening mechanism and the unified structured collaborative learning framework. Section V presents the experimental analysis and real-world deployment results. Finally, Section VI concludes the paper.

\section{RELATED WORK}
\subsection{Cooperative Navigation Using Reinforcement Learning}
Cooperative navigation using reinforcement learning focuses on enabling multiple agents to learn coordinated policies for efficient path finding and collision-free navigation in shared environments. Early studies\cite{12}, \cite{13}, \cite{14} establish learning-based formulations for multi-agent path finding and navigation, where reinforcement learning is combined with decentralized execution to enable scalable coordination. Graph-based methods\cite{15}, \cite{16} explicitly model inter-agent relationships and enable structured reasoning for navigation tasks, with graph neural networks and attention-based approaches improving performance in large-scale multi-agent path finding environments. Attention mechanisms\cite{17} are further explored to dynamically capture interaction relevance and improve decision making in complex navigation settings. In addition, communication learning methods\cite{18} allow agents to exchange task-relevant information, which enhances coordination efficiency in partially observable navigation tasks. Socially-aware navigation approaches\cite{19} incorporate interaction priors to ensure safe and efficient navigation in crowded environments. Hierarchical reinforcement learning\cite{20} and adaptive interaction modeling methods\cite{21}, \cite{22}, \cite{23} are also introduced to decompose complex tasks and capture dynamic topological relationships among agents. 

Despite these advances, most existing approaches rely on predefined interaction structures or lack mechanisms to incorporate structured prior knowledge into policy learning, limiting their adaptability in dynamic and uncertain environments. In contrast, the present study proposes a cooperative navigation framework that employs a knowledge-embedded attention as the actor to enhance representation learning under partial observability, and a GAT as the critic to explicitly model dynamic interaction topologies, thereby achieving improved coordination performance and robustness in complex multi-agent path finding tasks.

\subsection{Safe MARL}

Safe MARL focuses on learning cooperative policies that maximize performance while respecting safety constraints in dynamic and uncertain environments. A lagrangian-based extension\cite{24} improves scalability by converting safety constraints into adaptive penalty terms within policy optimization. Subsequent studies\cite{25} develop scalable constrained policy optimization methods that enable decentralized updates under local interactions while preserving safety guarantees in large-scale systems. To further enhance practical applicability, constraint projection approaches\cite{26} are introduced to enforce safety by directly correcting policy updates within feasible regions. Parallel to optimization-based methods, control-theoretic approaches\cite{27} integrate CBFs into multi-agent reinforcement learning to enforce safety constraints at the action level with formal guarantees. Robust extensions\cite{28} incorporate uncertainty estimation into CBFs to maintain safety under model mismatch and dynamic interactions. Recent work combines neural CBFs with attention mechanisms\cite{29} to address time-varying observability and disturbance robustness in safety-critical multi-agent systems. Furthermore, to guarantee safety, several MARL frameworks\cite{30} integrate Model Predictive Control (MPC) to avoid collision-inducing actions during training and deployment. 

Despite these advances, existing methods typically rely on abstract state representations or assume access to accurate system models, which limits their applicability in realistic perception-constrained environments. In contrast, the present study develops a safety-aware multi -agent reinforcement learning framework that integrates LiDAR-based perception with CBLFs to construct geometrically verifiable safety constraints, enabling reliable and physically consistent collision avoidance in cooperative navigation tasks.

\section{PROBLEM FORMULATION}

This paper considers a multi-agent system comprising $N$ differential-drive robots operating in a shared two-dimensional workspace. Each robot is subject to inherent underactuation constraints and actuator saturation, which strictly limit its feasible linear and angular velocities. Furthermore, each agent is equipped with a two-dimensional LiDAR sensor to acquire raw range measurements of the local environment for obstacle detection and state observation. Furthermore, the multi-agent system operates under limited communication capabilities, resulting in a time-varying interaction topology. Specifically, agents can exchange state information if and only if their relative distance is strictly less than a predefined maximum communication range $R_c$. Consequently, the number of observable neighbors fluctuates continuously, requiring the learning framework to robustly adapt to dynamic adjacency structures.

The primary objective of the cooperative navigation task is for all agents to navigate to their respective designated target positions within a predefined maximum time $T_{\max}$. During the navigation process, strict collision-free movements must be maintained. The task is evaluated as a failure if any collision occurs between an agent and static obstacles or peer agents. Additionally, if all agents fail to reach their target points before the time limit expires, the task is also considered a failure. Upon encountering any failure condition, the current episode is immediately terminated, and the entire environment along with the states of all agents are reset to initiate a new episode.

The kinematics of each differential-drive agent $i$ are formulated by the following continuous-time equations:

\begin{equation}
\label{e1}
\left\{\begin{aligned}\dot{x}_i &= v_i \cos \theta_i \\\dot{y}_i &= v_i \sin \theta_i \\\dot{\theta}_i &= \omega_i\end{aligned}\right.
\end{equation}
where $(x_i, y_i)$ denotes the two-dimensional cartesian position, and $\theta_i$ represents the heading angle. The control input is defined as $u_i = (v_i, \omega_i)$, corresponding to the linear and angular velocities. To account for actuator saturation, the control inputs are strictly bounded within predefined limits, such that $v_i \in [v_{\min}, v_{\max}]$ and $\omega_i \in [\omega_{\min}, \omega_{\max}]$. These non-holonomic constraints fundamentally restrict the lateral movement capabilities of the agents, thereby shaping the mathematically feasible action space for the cooperative navigation task.

Each agent is equipped with a LiDAR sensor to perceive the local environment. Due to the non-holonomic and underactuated constraints of the differential-drive robot, the sensor is configured to provide an azimuthal coverage of 180° directly in front of the vehicle. This field-of-view is discretized into 36 laser beams, yielding a 36-dimensional range measurement vector. 

\begin{figure*}[!t]
\centering
\subfloat[The architecture of the proposed navigation framework]{\includegraphics[width=4.2in]{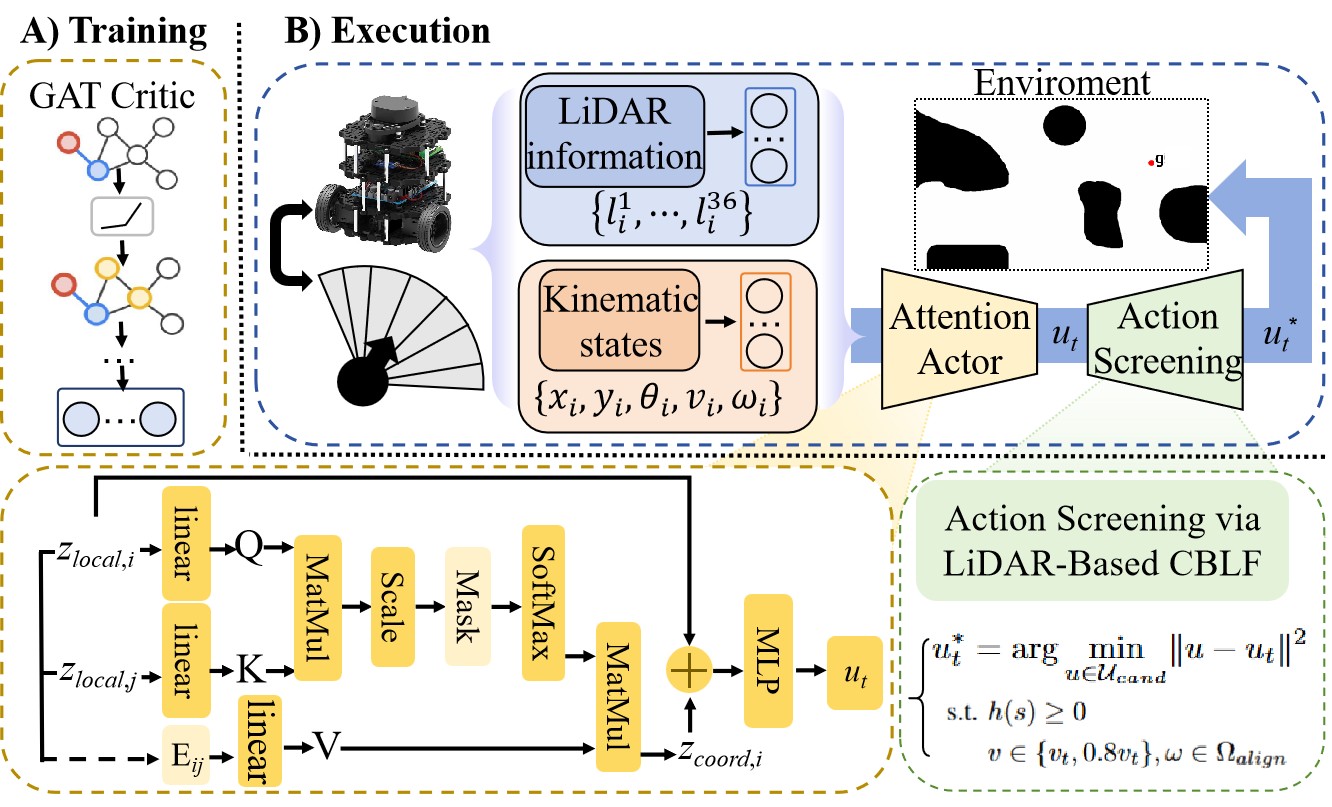}%
\label{fig1a}}
\subfloat[Time-varying topology]{\includegraphics[width=1.2in]{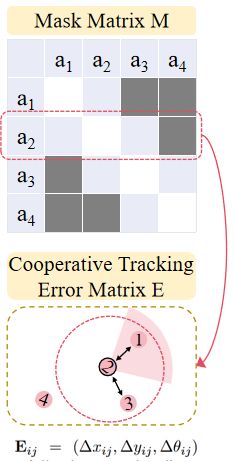}%
\label{fig1b}}
\caption{Overall architecture of the proposed navigation framework, comprising a centralized GAT-based critic for value estimation and a decoupled actor-screening pipeline where the attention-based actor generates coordination-aware actions $u_t$ via a value-reconstructed transformer (integrating tracking errors $E_{ij}$), followed by a LiDAR-based CBLF screening layer that maps $u_t$ to a geometrically safe control $u_t^*$ through discrete optimization.}
\label{fig1}
\end{figure*}

\section{THE GRAPH-BASED SAFE MULTI-AGENT REINFORCEMENT LEARNING}

As shown in Fig.\ref{fig1}, this section introduces a graph-based safe multi-agent reinforcement learning framework, which integrates a CBLF-based action screening layer with a structured coordination architecture comprising a attention-based actor and a GAT-based centralized critic.

\subsection{Multi-agent Reinforcement Learning Framework}
The multi-agent cooperative navigation task is formulated as a Partially Observable Multi-Agent Markov Decision Process (POMDP), defined by the tuple $\mathcal{M} = \langle \mathcal{N}, \mathcal{S}, \mathcal{U}, P, R, \Omega, \mathcal{O}, \gamma \rangle$. Here, $\mathcal{N} = \{1, \dots, N\}$ denotes the set of agents, $\mathcal{S}$ represents the global state space, and $\mathcal{U} = \prod_{i=1}^{N} \mathcal{U}_i$ is the joint action space, where each individual {\bf{action space}} $u_i = (v_i, \omega_i) \in \mathcal{U}_i$ is subject to actuator saturation and underactuation constraints. The communication topology is governed by a distance-dependent disk model with radius $R_c$: an undirected edge $(i,j)$ exists if and only if $\|p_i(t) - p_j(t)\| \leq R_c$. This proximity-based connectivity naturally induces a time-varying topology, restricting each agent $i$ to a local observation $o_i \in \Omega$ via $\mathcal{O}(s, i)$. This {\bf{observation space} }$o_i$ comprises the 36-dimensional LiDAR range measurements, the agent’s own kinematic state, and joint observations synthesized via knowledge embedding and attention mechanisms. The transition probability $P(s'|s, \mathbf{u})$ describes the environment dynamics, while $R = \{r_1, \dots, r_N\}$ denotes the reward functions. The objective of the multi-agent system is to find an optimal joint policy $\boldsymbol{\pi}$ that maximizes the expected discounted cumulative reward $J(\theta) = \mathbb{E}[\sum_{t=0}^{T_{\max}} \gamma^t r_i^t]$ while strictly satisfying the collision-free requirements within the time limit $T_{\max}$.

The total reward $r_i$ for agent $i$ at time $t$ is defined as a weighted sum of task completion, navigation efficiency, and motion constraints:

\begin{equation}
\label{2}
r_i = r_{task,i} + r_{nav,i} + r_{obs,i} + r_{motion,i}
\end{equation}

The task-related reward $r_{task,i}$ provides sparse feedback based on the terminal states of the agent:
\begin{equation}
\label{3}
r_{task,i} = \mathds{1}_{arrival} \cdot \beta_{A} + \mathds{1}_{collision} \cdot \beta_{C}
\end{equation}
where $\beta_{A}$ and $\beta_{C}$ represent the reward for reaching the target and the penalty for collision, respectively.

The navigation reward $r_{nav,i}$ incentivizes individual progress and collective convergence towards the targets:
\begin{equation}\label{4}
\begin{split}
r_{nav,i} = &w_0 \Delta d_{i,t} - w_1 + w_2 \Delta \bar{d}_t + w_3 \Delta d_{max,t} \\&+ \mathds{1}_{any\_arr} \cdot w_4 \Delta d_{i,t}
\end{split}
\end{equation}
where $\Delta d_{i,t}$ is the distance reduction of agent $i$, while $\Delta \bar{d}_t$ and $\Delta d_{max,t}$ denote the average and maximum distance reductions of the team.

The obstacle avoidance penalty $r_{obs,i}$ is formulated as a multi-level step function based on the minimum LiDAR range measurement $l_{min}$: 
\begin{equation}
\label{5}
r_{obs,i} = \begin{cases} \beta_1, & l_{min} < l_0 \\ \beta_2, & l_0 \le l_{min} < l_1 \\ \beta_3, & l_1\le l_{min} < l_2 \\ 0, & \text{otherwise} \end{cases}
\end{equation}
where $\beta_1 < \beta_2 < \beta_3 < 0$, and they are tiered obstacle avoidance penalty coefficients that increase as the minimum LiDAR range decreases.

The motion penalty $r_{motion,i}$ suppresses unstable rotation and state stagnation to ensure smooth trajectories:
\begin{equation}
\label{6}
r_{motion,i} = \beta_\omega |\omega_i| + \mathds{1}_{spin} \cdot \beta_{S} + \mathds{1}_{stagn} \cdot \beta_{st}
\end{equation}
where $\beta_{\omega}$ penalizes high angular velocity. Both $\beta_S$ and $\beta_{st}$ penalize the combination of low linear velocity and high angular velocity.

\subsection{Geometrically Verifiable Action Screening via LiDAR-based CBLF}

This section introduces a safety-decoupled action screening mechanism that maps the raw policy output $u_t$ into a geometrically feasible safe control input $u_t^*$. This process ensures strict collision avoidance by verifying the robot's future occupancy against the local geometric environment captured by the LiDAR.

\subsubsection{Kinematic Trajectory Prediction}

To evaluate the safety of a control input $u_t = (v_t, \omega_t)$, the robot’s pose $s_{t+n} = [x_{t+n}, y_{t+n}, \theta_{t+n}]^T$ after a prediction horizon $\Delta T = n \cdot \Delta t$ is computed. Based on the non-holonomic kinematics of the differential-drive robot, the trajectory is modeled as a circular arc:
\begin{equation}
\label{7}
s_{t+n} = \begin{cases} s_t + \begin{bmatrix} -\frac{v_t}{\omega_t} \sin\theta_t + \frac{v_t}{\omega_t} \sin(\theta_t + \omega_t \Delta T) \\ \frac{v_t}{\omega_t} \cos\theta_t - \frac{v_t}{\omega_t} \cos(\theta_t + \omega_t \Delta T) \\ \omega_t \Delta T \end{bmatrix}, & |\omega_t| \geq \epsilon \\ s_t + \begin{bmatrix} v_t \Delta T \cos\theta_t \\ v_t \Delta T \sin\theta_t \\ 0 \end{bmatrix}, & |\omega_t| < \epsilon \end{cases}
\end{equation}
where $\epsilon = 0.01$ is a small threshold to prevent numerical instability. This prediction enables the assessment of potential spatial occupancy before the control command is executed.

\subsubsection{Geometrically Verifiable Safety Function}

This paper defines a CBLF $h(s)$ based on the geometric coverage of LiDAR visible domains. 

The function $h(s)$ is designated as a CBLF rather than a standard CBF to accommodate the discrete nature of LiDAR-based spatial sampling and the non-differentiability of the min-projection operator. The proposed LiDAR-based CBLF operates as a rigorous one-step discrete safety filter. It strictly guarantees that the subsequent discrete state $s_{t+n}$, resides within the geometrically verifiable safe set $\mathcal{C}$ given the condition $h(s) \ge 0$.

Let $\mathcal{C}(s_{t+n}, r_{safe})$ be a safety disk centered at the predicted position. The safety function $h(s)$ determines the distance between the disk boundary and the perceived environmental constraints:

\begin{figure}[!t]
\centering
\includegraphics[width=3.5in]{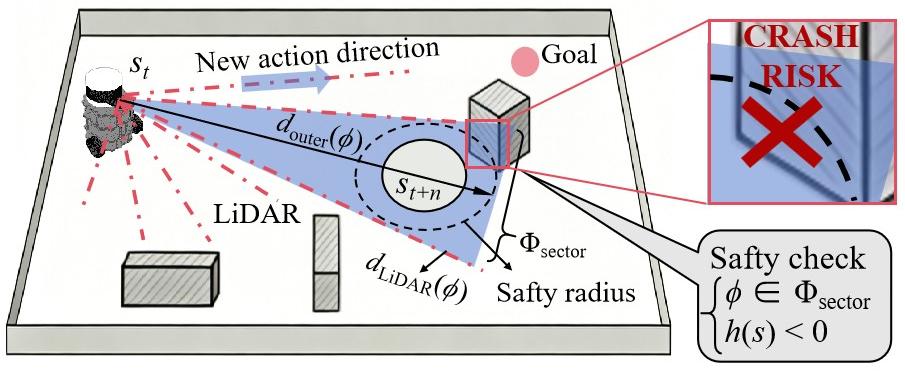}%
\caption{This diagram intuitively illustrates the geometric principles of the LiDAR-based CBLF safe passageway. It constructs a safe sector around the predicted state $s_{t+n}$ that must accommodate the required safety radius $r_{\text{safe}}$. The constraint $h(s) \ge 0$ is satisfied when this entire blue sector remains within the boundaries detected by the raw LiDAR rays.}
\label{fig11}
\end{figure}

\begin{equation}
\label{8}
h(s) \!=\! \min \!\!\left(\!\! R_{\mathrm{max}}\!-\!d_{\mathrm{outer}},\!\!\!\min_{\phi \in \Phi_{\mathrm{sector}}}\!\!\!\!\!\! \{d_{\mathrm{LiDAR}}(\phi)\!-\!d_{\mathrm{outer}}(\phi)\} \!\!\right)
\end{equation}
where $R_{\max}$ denotes the maximum effective range of the LiDAR sensor. As shown in Fig.\ref{fig11}, $d_{LiDAR}(\phi)$ is the 36-dimensional range measurement, $\Phi_{\mathrm{sector}}$ is the angular interval covered by the safety disk in the ego-coordinate system, and $d_{\mathrm{outer}}$ is the maximum distance from the current robot center to the boundary of the safety disk. A state is considered geometrically safe if $h(s) \geq 0$, ensuring the safety disk is entirely within the sensor's valid field-of-view and clear of any detected obstacles.

\subsubsection{Discrete Safety-Decoupled Optimization}

Unlike traditional continuous Quadratic Programming (QP), which may struggle with raw sensor-domain constraints, this paper employs a discrete search-based optimization to find the safe action $u_t^*$. If the initial policy output $u_t$ results in $h(s) < 0$, the screening mechanism explores a discrete candidate set $\mathcal{U}_{cand}$:

\begin{equation}
\label{8}
\begin{aligned} u_t^* = \arg\min_{u \in \mathcal{U}_{cand}} & \|u - u_t\|^2 \\ \text{s.t. } & h(s) \geq 0 \\ & v \in \{v_t, 0.8v_t\}, \omega \in \Omega_{align} \end{aligned}
\end{equation}

The candidate set is constructed by discretizing the linear velocity and searching for angular velocities $\omega$ that align the predicted trajectory with the safe LiDAR beams. By iteratively expanding the search from the nearest beam, the mechanism guarantees a physically consistent correction that respects the actuator limits while prioritizing safety.

\subsection{Attention-based Actor Design}

The actor network integrates a local feature extraction pathway with a restructured attention module. For each agent $i$, the input vector $\zeta_i = [l_i^1, \dots, l_i^{36}, x_i, y_i, \theta_i, v_i, \omega_i]^T$ combines 36-dimensional LiDAR ranges and 5-dimensional kinematic states. A linear projection maps these observations into a compact local representation $z_{local,i} = \text{Linear}(\zeta_i)$, characterizing individual controllability and obstacle avoidance constraints.

As shown in Fig.\ref{fig1b}, to model dynamic neighborhood interactions, this paper proposes a value-reconstructed attention where the attention mechanism is conditioned on collaborative tracking errors. Unlike standard self-attention where $Q, K, V$ are derived from the same embedding via linear projections, this paper defines $V$ as the cooperative tracking error tensor $\mathbf{E} \in \mathbb{R}^{N \times N \times 3}$, where each element $\mathbf{E}_{ij} = (\Delta x_{ij}, \Delta y_{ij}, \Delta \theta_{ij})$ represents the relative pose. Crucially, because the diagonal of this error matrix consists entirely of zeros ($\mathbf{E}_{ii}=0$), it lacks the requisite feature variance for identity mapping and self-attention, meaning it cannot be used directly as the standard input for $Q, K,$ and $V$ simultaneously. Consequently, the query $\mathbf{Q}_i = W^Q z_{local,i}$ and key $\mathbf{K}_j = W^K z_{local,j}$ are projected from state embeddings to determine interaction priorities, while the value $\mathbf{V}_{ij} = W^V \mathbf{E}_{ij}$ undergoes an analogous transformation to inject explicit geometric and physical guidance into the coordination process.

To accommodate time-varying communication topologies within the sensing range $R_c$, a masking matrix $\mathbf{M}$ is integrated into the encoder structure, defined as:

\begin{equation}
\label{9}
\mathbf{M}_{ij} = \begin{cases} 0, & \text{if } \|p_i - p_j\| \leq R_c \\ -\infty, & \text{otherwise} \end{cases}
\end{equation}

The structural feature $z_{coord,i}$ is then computed via the attention mechanism:

\begin{equation}
\label{10}
\text{Attn}(\mathbf{Q, K, V}) = \text{softmax}(\frac{\mathbf{QK}^T}{\sqrt{d_k}} + \mathbf{M}) \mathbf{V}.
\end{equation}

The final policy combines these features through concatenation to sample the control action $u_i \sim \mu_{\theta}(u_i | z_{local,i} \Vert  z_{coord,i})$. $\Vert$ denotes the concatenation operator. This dual-pathway architecture ensures that the policy remains scalable and robust across varying agent counts while maintaining high-precision coordination in dynamic environments.

\subsection{GAT-based Critic}

To overcome the scalability and interaction modeling limitations of independent MLP critics, this paper proposes a centralized critic based on GAT. Each agent is modeled as a node $i$ in a dynamic graph with a 44-dimensional feature vector $\mathbf{h}_i$, which integrates 36-dimensional LiDAR data, 5-dimensional self-state $(x_i, y_i, \theta_i, v_i, \omega_i)$, and 3-dimensional relative goal information$(\Delta x_{ig}, \Delta y_{ig}, \Delta \theta_{ig})$. The GAT layer adaptively captures the inter-agent influence by computing the attention coefficient $\omega_{ij}$ and aggregating neighbor features through the following scoring and transformation process:

\begin{equation}
\label{11}
\left\{\begin{aligned}\pi_{ij} &= \mathbf{v}_{\text{att}}^T \text{LeakyReLU}\left([\mathbf{W}\mathbf{h}_i \, \Vert \, \mathbf{W}\mathbf{h}_j]\right) \\\omega_{ij} &= \frac{\exp(\pi_{ij})}{\sum_{k \in \mathcal{N}_i} \exp(\pi_{ik})} \\\mathbf{h}'_i &= \sigma \left( \sum_{j \in \mathcal{N}_i} \omega_{ij} \mathbf{W} \mathbf{h}_j \right)\end{aligned}\right.
\end{equation}
where $\mathbf{h}_i, \mathbf{h}_j \in \mathbb{R}^p$ denote the input feature vectors of agent $i$ and its neighbor $j$; $\mathbf{W} \in \mathbb{R}^{q \times p}$ is a shared learnable linear transformation matrix that projects the raw state features into a higher-dimensional representational space $q$, $\mathbf{v}_{\text{att}} \in \mathbb{R}^{2q}$ is a learnable attention parameter vector, which maps the concatenated features into a scalar raw attention score $\pi_{ij} \in \mathbb{R}$. And $\mathcal{N}_i$ represents the set of neighbors within the communication range $R_c$. By utilizing multi-head attention and shared GAT parameters, the critic effectively filters task-relevant structural information from the swarm, outputting a value estimate vector for all agents to facilitate accurate global credit assignment during centralized training.

\begin{figure*}[!t]
\centering
\subfloat[Comparison study]{\includegraphics[width=1.6in]{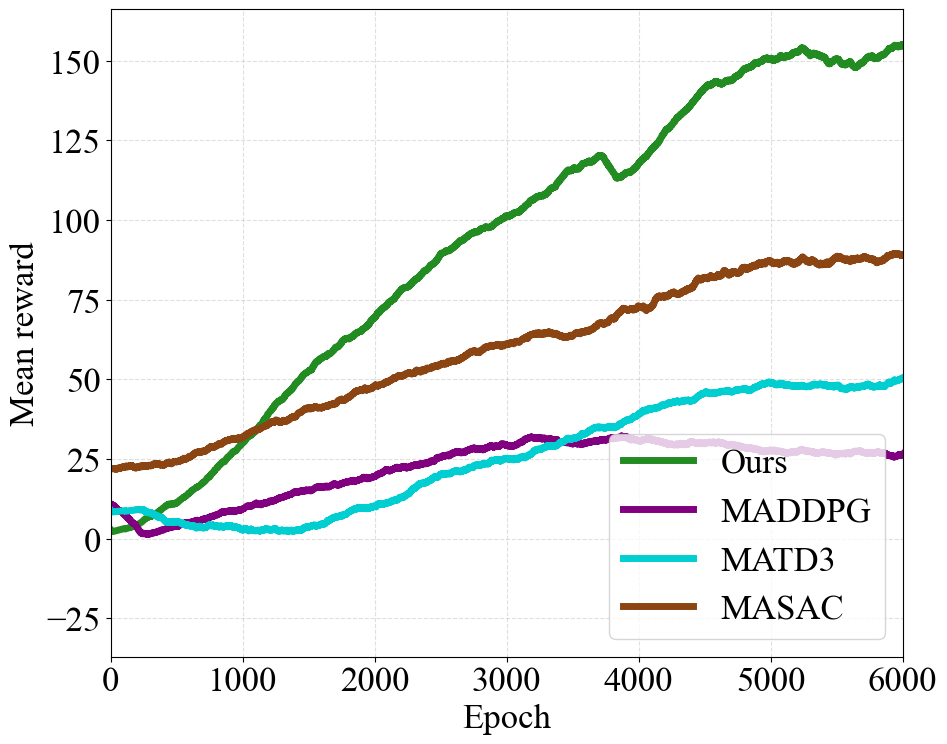}%
\label{fig2a}}
\hfil
\subfloat[Impact of Action Screening]{\includegraphics[width=1.6in]{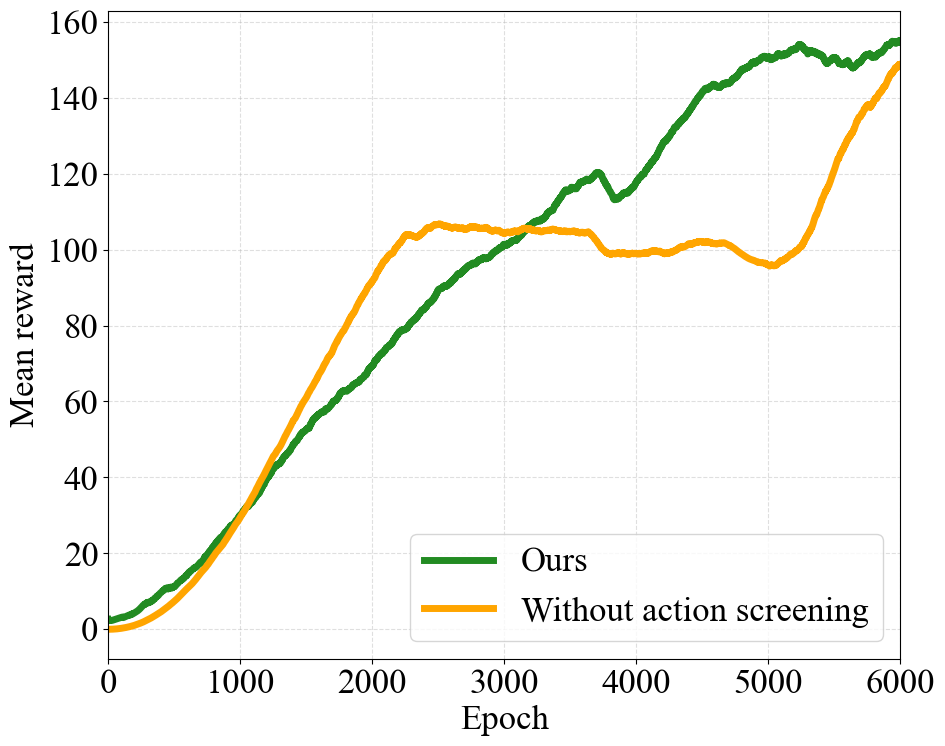}%
\label{fig2b}}
\hfil
\subfloat[Impact of Attention-based Actor]{\includegraphics[width=1.6in]{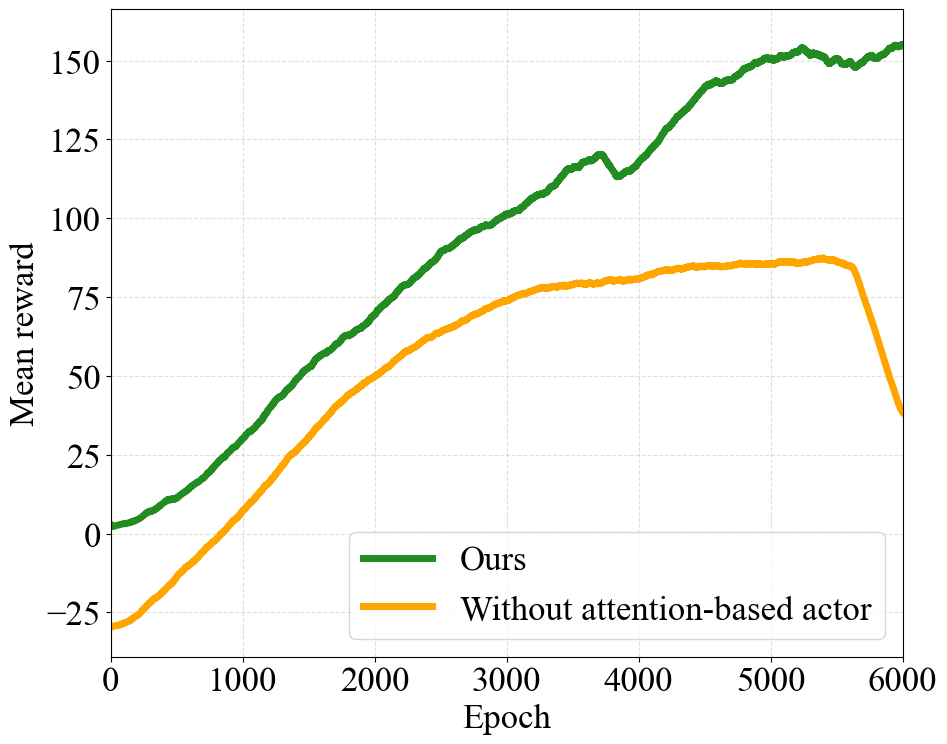}%
\label{fig2c}}
\hfil
\subfloat[Impact of GAT-based Critic]{\includegraphics[width=1.6in]{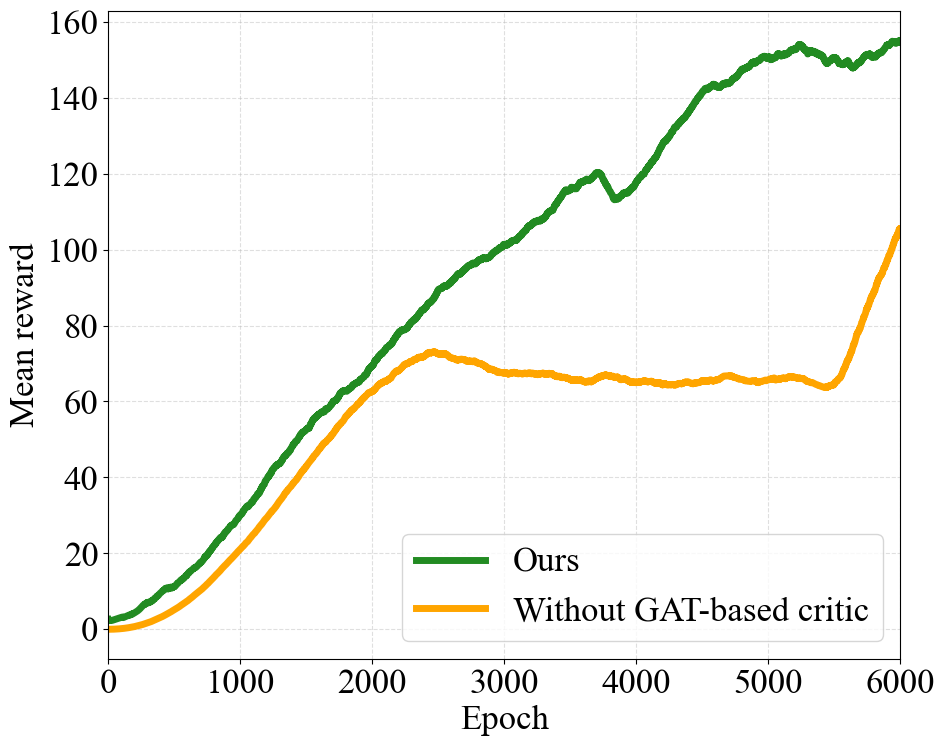}%
\label{fig2d}}

\caption{Learning curves of the mean reward across 6000 training epochs. The solid lines represent the average reward, showcasing the sample efficiency, convergence speed, and final policy performance of the proposed method compared to baselines and its own ablation variants.}
\label{fig2}
\end{figure*}

\section{SIMULATION AND EXPERIMENTAL RESULTS}

This chapter evaluates the performance of the proposed algorithm through comprehensive simulation and physical experiments. The Multi-Agent Proximal Policy Optimization (MAPPO\cite{32}) serves as the baseline algorithm. All training and simulation tasks are conducted on a workstation equipped with an AMD Ryzen 9 7950X CPU and an NVIDIA GeForce RTX 3070 GPU. The simulation environment is developed using the ir-sim platform\cite{31}, where the virtual agents are modeled to strictly retain the critical kinematic constraints and sensor configurations of the Turtlebot3 platform, including its non-holonomic drive limits and LiDAR sensing characteristics. For real-world validation, an experimental arena is configured with four Turtlebot3 mobile robots to demonstrate the algorithm's sim-to-real transferability and coordination efficiency in physical environments.

\subsection{Evaluation Metrics and Experimental Design}

To quantitatively evaluate the performance of the proposed algorithm and its variants, all experiments are conducted over 100 independent test trials. A trial is defined as a failure if any agent fails to reach its target within the maximum allowed time steps or experiences a collision during transit. The following three metrics are utilized: Success Rate (SR), Collision Rate (CR) and Average Steps (AS).

In the training phase, a collision triggers an immediate environment reset. Consequently, while the agents acquire basic obstacle avoidance, they often lack the robustness to handle sudden obstacles. A critical challenge arises when a leading agent reaches its goal and stops, effectively becoming a static obstacle for trailing agents. To analyze these factors, this paper conducts:
\begin{enumerate}
\item \textbf{Comparative Study}: Testing the impact of three different arrival distances on algorithm performance to evaluate sensitivity to goal proximity.
\item \textbf{Ablation Study}: Investigating the individual contributions of the proposed action screening (CBLF and LiDAR-based), the attention-based actor, and the GAT-based critic to the overall safety and coordination efficiency of the swarm.
\item \textbf{Real-World Experiments}: Validating the sim-to-real transferability and practical robustness of the proposed framework using four TurtleBot3 robots. These experiments assess the system's performance in handling sensor noise and unmodeled physical dynamics while ensuring rigorous collision avoidance in a physical indoor arena.
\end{enumerate}

\begin{table*}[t]
\centering
\caption{THIS TABLE SUMMARIZES THE PERFORMANCE OF The APPROACH AGAINST MASAC, MATD3, AND MADDPG ACROSS THREE ARRIVAL DISTANCE CONFIGURATIONS, ALONGSIDE AN ABLATION STUDY OF THE CORE COMPONENTS.}
\label{table1}
\resizebox{\textwidth}{!}{
\begin{tabular}{|c|c|c|c|c|c|c|c|c|c|}
\hline
 & \multicolumn{3}{c|}{\textbf{Arrival Distance = $1m$}} & \multicolumn{3}{c|}{\textbf{Arrival Distance = $0.8m$}} & \multicolumn{3}{c|}{\textbf{Arrival Distance = $0.3m$}} \\ \hline
\textbf{Agent} & \textbf{Success $\uparrow$} & \textbf{Collisions $\downarrow$} & \textbf{Steps $\downarrow$} & \textbf{Success $\uparrow$} & \textbf{Collisions $\downarrow$} & \textbf{Steps $\downarrow$} & \textbf{Success $\uparrow$} & \textbf{Collisions $\downarrow$} & \textbf{Steps $\downarrow$} \\ \hline
\textbf{Ours (Full)} & \textbf{99\%} & \textbf{0\%} & \textbf{125} & \textbf{98\%} & \textbf{0\%} & \textbf{140} & \textbf{95\%} & \textbf{0\%} & \textbf{162} \\ \hline
MASAC\cite{35} & 78\% & 8\% & 138 & 72\% & 12\% & 155 & 65\% & 18\% & 178 \\ \hline
MATD3\cite{34} & 72\% & 18\% & 152 & 55\% & 24\% & 172 & 48\% & 32\% & 195 \\ \hline
MADDPG\cite{33} & 12\% & 75\% & 210 & 8\% & 82\% & 235 & 5\% & 91\% & 250 \\ \hline
\multicolumn{10}{|c|}{\textbf{Ablation Study: Effectiveness of Proposed Components}} \\ \hline
Ours w/o CBLF Screening & 68\% & 31\% & 148 & 61\% & 32\% & 142 & 52\% & 45\% & 160 \\ \hline
Ours w/o Attention Actor & 82\% & 0\% & 155 & 75\% & 0\% & 175 & 70\% & 0\% & 198 \\ \hline
Ours w/o GAT Critic & 84\% & 0\% & 145 & 78\% & 0\% & 168 & 72\% & 0\% & 185 \\ \hline
\end{tabular}
}
\end{table*}

\subsection{Comparative Result Analysis}
As shown in Fig. 2a and Table \ref{table1}, the proposed method significantly outperforms MADDPG, MATD3, and MASAC in both training efficiency and final navigation performance. From a learning perspective, the reward curve of the proposed framework exhibits a rapid ascent during the early epochs, reaching a superior reward ceiling that far exceeds all baselines. Quantitatively, the approach is the only one to maintain a perfect zero-collision record across all arrival distance configurations. In table \ref{table1}, the percentages and values reflect the average results over 100 independent trials for each scenario. The ablation study further highlights that disabling the CBLF-based screening layer leads to a surge in collision rates and a substantial decline in success, particularly in high-proximity settings. While omitting the attention actor or GAT critic does not compromise safety (as CBLF remains active), it results in increased average steps, demonstrating their critical role in optimizing coordination efficiency.


\begin{figure*}[!t]
\centering
\subfloat[Initial state]{\includegraphics[width=1.6in]{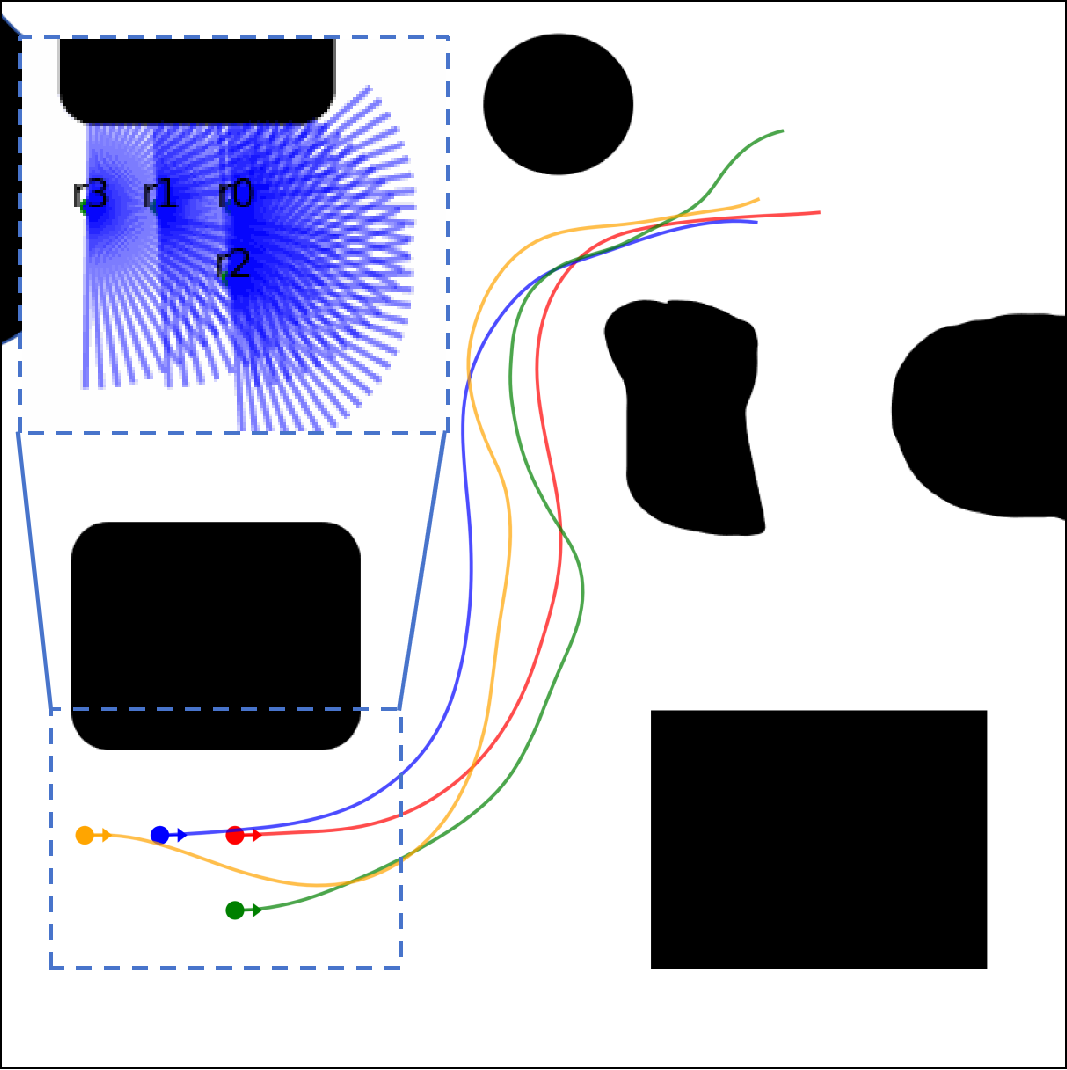}%
\label{fig3a}}
\hfil
\subfloat[Obstacle avoidance in navigation]{\includegraphics[width=1.6in]{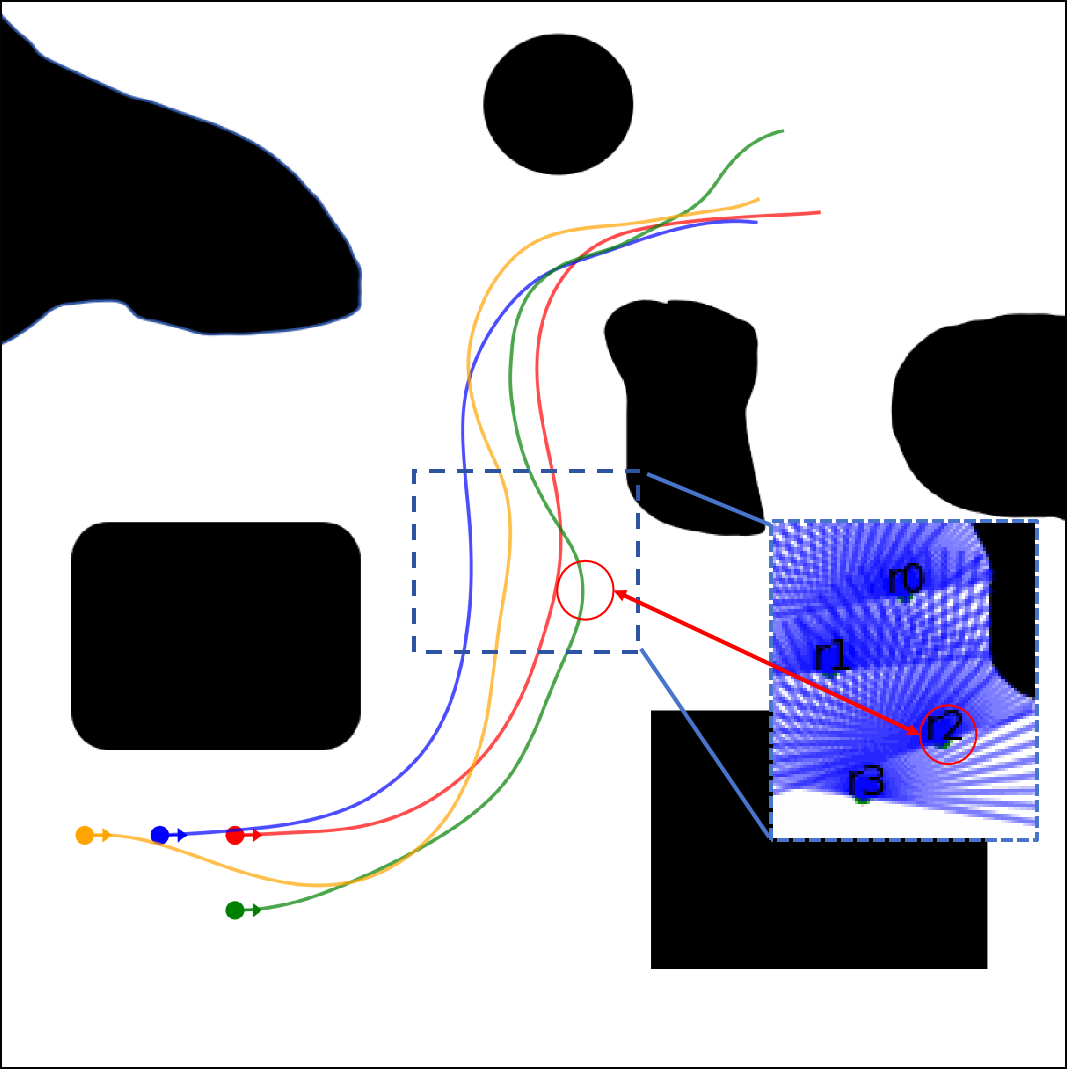}%
\label{fig3b}}
\hfil
\subfloat[Emergence of sudden obstacle ]{\includegraphics[width=1.6in]{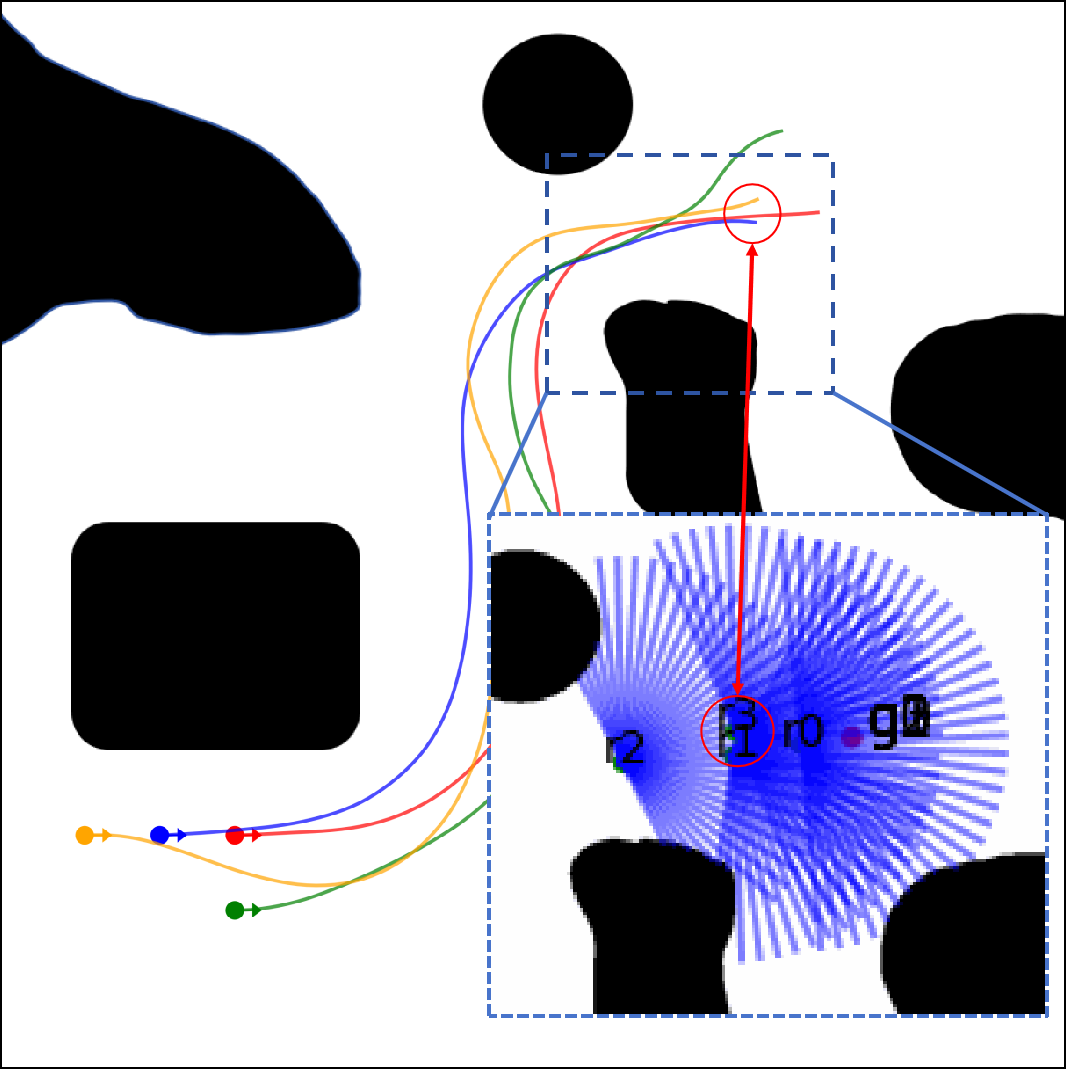}%
\label{fig3c}}
\hfil
\subfloat[Safety intervention]{\includegraphics[width=1.6in]{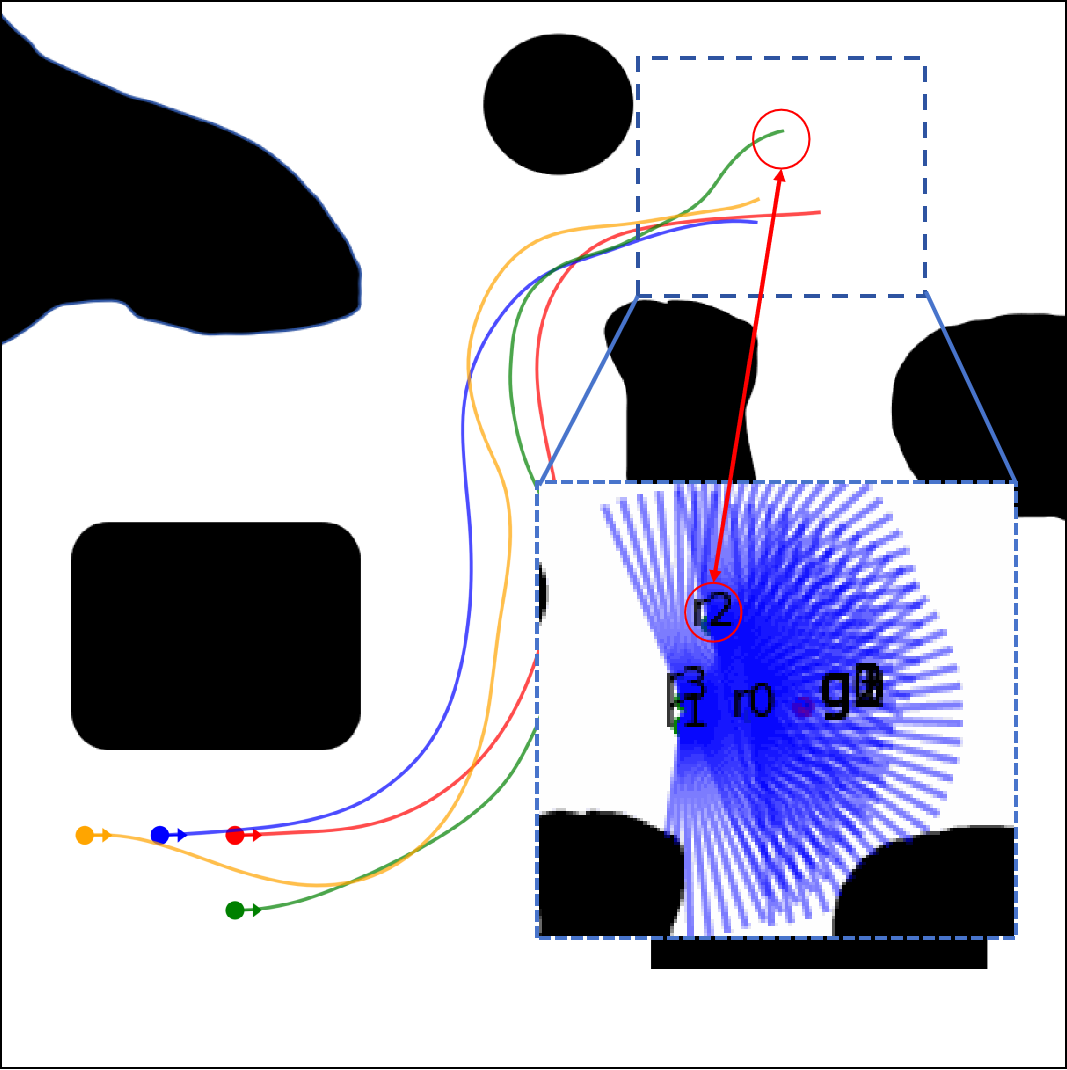}%
\label{fig3d}}

\caption{These figures illustrates the sequential snapshots of the multi-agent navigation task. Starting from the initial configuration (A), the agents maintain cooperative trajectories during the execution phase (B). The scenario highlights a critical challenge where the leading agents stop upon reaching their goals, forming sudden obstacles for the trailing swarm (C). The proposed CBLF-based action screening mechanism successfully identifies the collision risk and corrects the agents' velocities to ensure a smooth and collision-free path to the remaining targets (D).}
\label{fig3}
\end{figure*}

As shown in Fig.\ref{fig3}, the trajectory analysis provides direct evidence of the necessity and efficacy of the decoupled safety architecture. During the initial and intermediate stages, the attention-based actor generates efficient coordination behaviors, enabling the swarm to navigate through dynamic topologies. 

As shown in Fig.\ref{fig3c}, the sudden stoppage of preceding agents creates high-risk constraints that were not previously encountered during policy exploration. The CBLF action screening layer activates in real-time. By projecting the nominal control inputs into a geometrically verifiable safe domain, the agents execute precise evasive maneuvers as depicted in Fig.\ref{fig3d}. This intervention ensures that the minimum safety radius is never violated, effectively bridging the gap between the learned coordination strategy and the rigid requirements of physical safety.

\begin{table}[h]
\centering
\caption{Performance of the proposed approach with varying agent numbers ($N$) at a fixed target distance of $1m$.}
\label{table2}
\begin{tabular}{|c|c|c|c|}
\hline
\textbf{No. of Agents} & \textbf{Success $\uparrow$} & \textbf{Collisions $\downarrow$} & \textbf{Timeouts $\uparrow$} \\ \hline
5 & 88\% & 0\% & 12\% \\ \hline
6 & 84\% & 0\% & 16\% \\ \hline
7 & 78\% & 0\% & 22\% \\ \hline
8 & 68\% & 0\% & 32\% \\ \hline
\end{tabular}
\end{table}

As shown in Table \ref{table2}, the scalability analysis reveals a distinct trade-off between absolute safety and operational efficiency in dense swarms. While the proposed approach maintains a zero-collision record across all configurations, the success rate declines from 88.0\% to 68\% as $N$ increases. This performance drop is primarily driven by spatial congestion near the target area. As leading agents occupy the limited space around the goals, the CBLF-based safety layer prevents trailing agents from forcing entry into these high-risk, crowded regions. Consequently, high agent density forces these agents to wander or wait outside the congested zone, eventually leading to mission failure due to timeouts.

\subsection{Ablation Study Analysis}

To provide a more comprehensive validation of the proposed architecture, this section delivers an extended analysis linking the quantitative data from Table \ref{table1} with the qualitative trends observed in the learning curves Fig.\ref{fig2}. The ablation study dissects the contribution of the CBLF screening layer, the attention-based actor, and the GAT-based critic.

\subsubsection{CBLF Screening Layer} 

As shown in Table \ref{table1}, the version "w/o CBLF Screening" suffers from high collision rates, reaching 45\% at the $0.3m$ arrival distance. As shown in Fig.\ref{fig2a}, while the agent occasionally achieves high rewards through aggressive maneuvers, the performance is characterized by extreme volatility. Without the geometric constraints of the CBLF, the agent cannot distinguish between high-reward coordination and lethal collision risks. The CBLF acts as a physical inductive bias, filtering out unsafe actions and allowing the policy to converge on an optimal solution within a protected manifold.

\subsubsection{Attention-based Actor} 

Removing the attention module results in a significant increase in average steps. The learning curve in Fig.\ref{fig2b} shows a stagnant reward ceiling and severe instability after 6000 epochs. This confirms that the value-reconstructed attention, which utilizes the cooperative tracking error matrix $E_{ij}$, is critical for capturing the spatial sensitivity required for precision navigation. Without it, the model relies on simple MLP feature extraction, which fails to generalize across the dynamic topological changes inherent in swarm systems.

\subsubsection{GAT-based Critic}

The impact of the GAT-based critic is most evident in the convergence speed and stability. As shown in Fig.\ref{fig2c}, the version without the GAT critic has a substantially flatter learning curve, failing to reach the high-reward regime within the same training duration. Quantitatively, Table \ref{table1} reveals a drop in success rate even at three kinds of arrival distance, suggesting that the agents struggle to understand their individual contributions to the group success. The GAT architecture provides a superior value estimation by adaptively weighing neighboring influences, which reduces the gradient variance in the MAPPO update and leads to a more robust policy.

\subsection{Real World Experiments}

To validate the sim-to-real transferability and the practical robustness of the proposed framework, this paper conducted physical experiments in an indoor arena using four TurtleBot3 robots. Each robot is a differential-drive platform equipped with a 360° 2D LiDAR. Furthermore, each robot is integrated with an Inertial Measurement Unit (IMU) and wheel encoders for odometry. By deploying the robots from known fixed initial positions, precise real-time state estimation—including both position and orientation—is achieved through the fusion of these onboard sensors.

\begin{figure}[H]
\centering
\subfloat[Initial position]{\includegraphics[width=1.6in,height=1.4in]{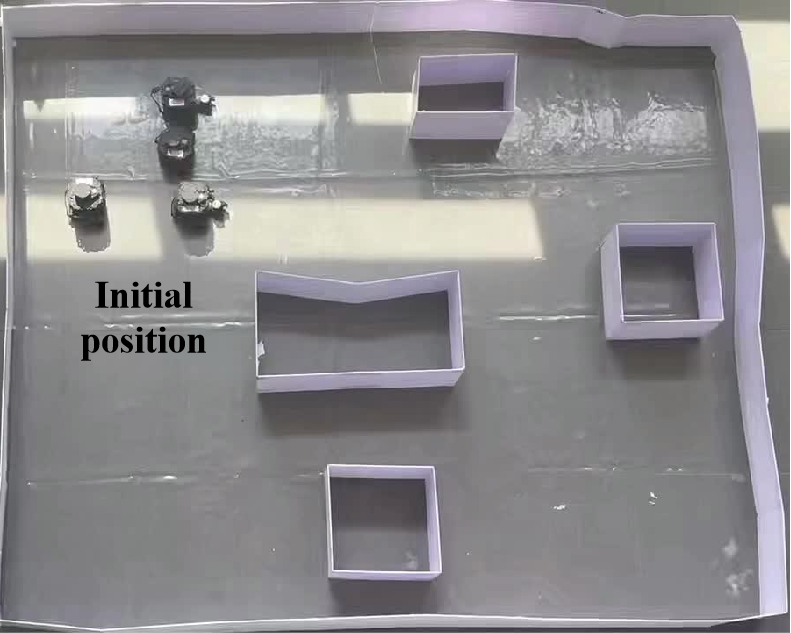}%
\label{fig4a}}
\hfil
\subfloat[Across narrow corridor]{\includegraphics[width=1.6in,height=1.4in]{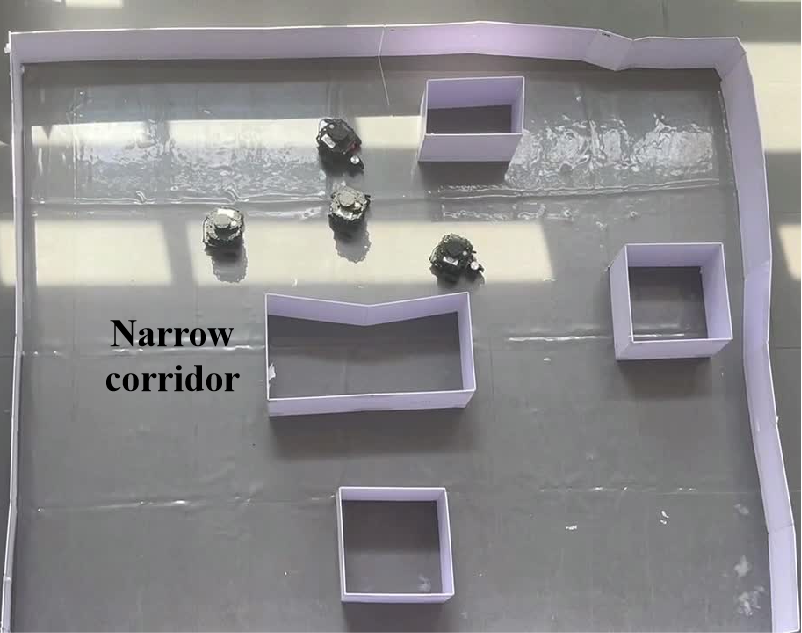}%
\label{fig4b}}
\hfil
\subfloat[Obstacle avoidance]{\includegraphics[width=1.6in,height=1.4in]{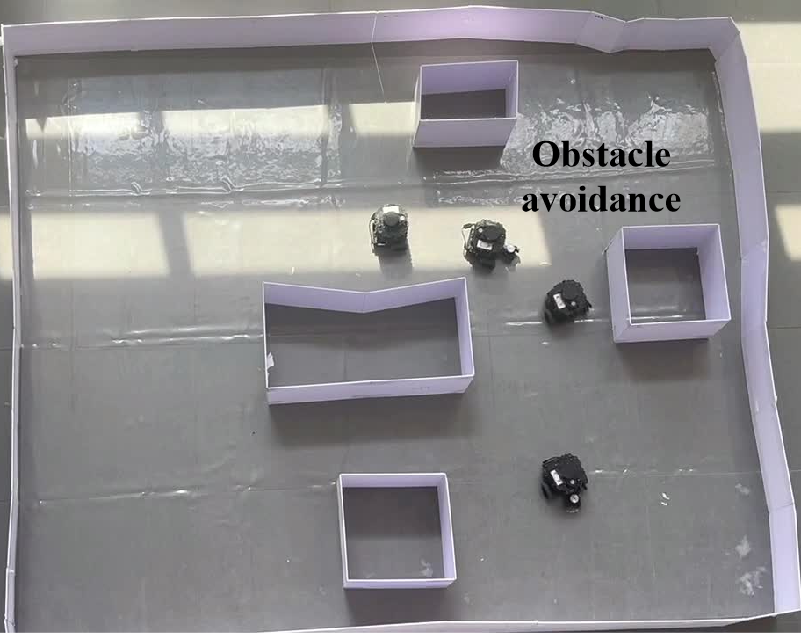}%
\label{fig4c}}
\hfil
\subfloat[Reach the goal]{\includegraphics[width=1.6in,height=1.4in]{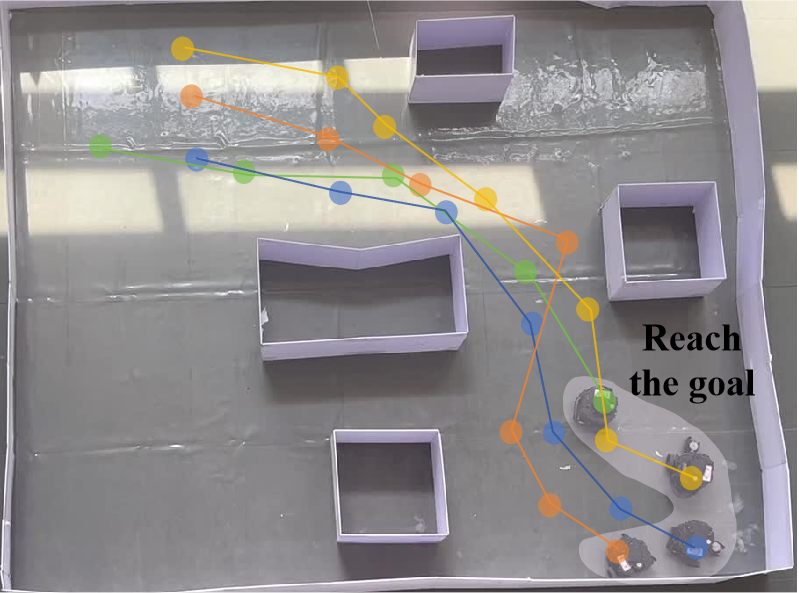}%
\label{fig4d}}

\caption{This figure illustrates the real-world validation of the navigation framework. The results demonstrate that the robots successfully reach their targets while maintaining a safe inter-agent distance throughout the process.}
\label{fig4}
\end{figure}

As shown in Fig.\ref{fig4}, the attention-based actor successfully manages the dense topological interactions between the four robots, generating smooth paths that avoid the oscillatory behaviors often seen in decentralized RL. More importantly, the CBLF action screening layer provides an essential safety buffer. It compensates for the inherent variance in LiDAR sampling and ensures that no physical collisions occur, even when the robots maneuver in close proximity during the mid-stage crossover. This successful deployment proves that the geometrically verifiable safety function effectively bridges the gap between simulated safety and real-world execution.

\section{CONCLUSION}

This paper presents a multi-agent reinforcement learning framework designed for coordinated navigation under non-holonomic constraints and dynamic topologies. By introducing the geometrically verifiable control barrier-like function (CBLF), this paper achieved a robust action screening mechanism that guarantees zero collisions throughout both the learning and execution phases. The proposed value-reconstructed attention-based actor, enhanced by cooperative tracking error matrices, provides explicit geometric guidance for stable swarm coordination. The GAT-based critic ensures accurate global credit assignment. Experimental evaluations in both sim-to-real TurtleBot3 scenarios and simulations demonstrate that the proposed approach significantly outperforms competitive baselines like MASAC and MATD3, particularly in the environments with sudden obstacles. Future research will focus on addressing large-scale swarm navigation in highly uncertain dynamic environments.

\bibliographystyle{unsrt}
\bibliography{ref_RAL_abbrev}

\vfill
\end{document}